\documentclass[twocolumn]{article}
\usepackage[utf8]{inputenc}
\usepackage{fullpage}
\usepackage{graphicx,pgf,tikz} 
\usepackage{amsmath,amssymb,amsfonts,amsthm} 
\usepackage{multirow,multicol,booktabs} 
\usepackage{enumitem} 
\usepackage{lipsum,mwe} 
\usepackage{url}
\usepackage{times}

\definecolor{mypurple}{HTML}{6457A6}
\definecolor{myyellow}{HTML}{DBD56E}
\definecolor{mybrown}{HTML}{664E4C}
\definecolor{myred}{HTML}{8C271E}
\definecolor{mygreen}{HTML}{A4C2A8}
\definecolor{myblack}{HTML}{000000}

\newcommand{\bemph}[1]{\textbf{#1}}

\title{Discrete and Continuous Deep Residual Learning Over Graphs}
\author{
    Pedro H.C. Avelar\footnote{Universidade Federal do Rio Grande do Sul,
Brazil, emails: \{phcavelar,artavares,lamb\}@inf.ufrgs.br}
    \and Anderson R. Tavares\footnotemark[1]
    \and Marco Gori\footnote{Università Degli Studi di Siena,
Italy, email: marco@unisi.it}
    \and Luis C. Lamb\footnotemark[1]
}

\date{}

\begin{document}

\maketitle

\begin{abstract}
    In this paper we propose the use of continuous residual modules for graph kernels in Graph Neural Networks. We show how both discrete and continuous residual layers allow for more robust training, being that continuous residual layers are those which are applied by integrating through an Ordinary Differential Equation (ODE) solver to produce their output. We experimentally show that these residuals achieve better results than the ones with non-residual modules when multiple layers are used, mitigating the low-pass filtering effect of GCN-based models. Finally, we apply and analyse the behaviour of these techniques and give pointers to how this technique can be useful in other domains by allowing more predictable behaviour under dynamic times of computation.
\end{abstract}

\section{Introduction}

Graph Neural Networks (GNNs) are a promising framework to combine deep learning models and symbolic reasoning. 
Whereas conventional deep learning models, such as Convolutional Neural Networks (CNNs), effectively handle data represented in euclidean space, such as images, GNNs  generalise their capabilities to handle non-Euclidean data, such as relational data with complex relationships and interdependencies between entities. 

Recently, deep learning techniques such as pooling, dynamic times of computation, attention, and adversarial training, which  advanced the state-of-the-art in conventional deep learning (e.g. in CNNs), have been investigated in GNNs as well \cite{battaglia2018gn,kipf2017gcn,velickovic2018gat,xu2018powerfulgnn}.
Discrete residual modules, whose learned kernels are discrete derivatives over their inputs, have been proven effective to improve convergence and reduce the parameter space on CNNs, surpassing the state-of-the-art in image classification and other applications \cite{he2016resnet}. Given their effectiveness, the technique has been applied in many different areas and meta-models of deep learning to improve convergence and reduce the parameter space. Unfortunately, it has been shown that Graph Neural Networks (GNN) often ``fail to go deeper'' \cite{li2018gcndeeper,wu2019comprehensive}, with some work already arguing for residual connections \cite{kipf2017gcn} to alleviate, but not solve, this issue.

Further, there has been recent work in producing continuous residual modules \cite{chen2018odenet} that are integrated through Ordinary Differential Equation (ODE) Solvers. They have shown how these models can be used to substitute both recurrent and convolution-residual modules for small problems such as recognising digits from the MNIST dataset, regressing a trajectory and generating spirals from latent data. Further work has already been done exploring generative models \cite{gratwohl2018odenet}, using adversarial training for generating both data from synthetic distributions as well as producing high-quality samples from the MNIST and CIFAR-10 datasets.

In this paper we investigate the use of both discrete and continuous residual modules in learning kernels that operate on relational data, providing improvements over their non-residual counterparts, as well as a comparative analysis of the benefits and issues of applying these techniques to graph-structured data. The remainder of this paper is organised as follows: \textbf{Section~\ref{sec:gnn}} presents a brief survey on Deep Learning models and formalisations for relational data -- which we amalgamate under the GNN framework. In \textbf{Section~\ref{sec:res}}, we provide information on how to rework graph-based kernels into residual modules, to be used in the context of continuous-residual modules, and discuss their possible advantages and disadvantages. In \textbf{Section~\ref{sec:exp}}, we provide the experimental results we collected from converting graph modules to work residually and compare them to their non-residual counterparts. Finally, in \textbf{Sections \ref{sec:dis},  \ref{sec:relwk} and  \ref{sec:concfut}} we interpret the results, discuss related work, and point out directions for future research.

\section{On Graph Neural Networks}\label{sec:gnn}

In this section we describe the basics of well-known Graph Neural Network models. We do so by presenting some models which have been widely used in recent applications. For more comprehensive reviews of the field, please see e.g. \cite{battaglia2018gn,gilmer2017mpnn,wu2019comprehensive}.

One of the first formalisations of GNNs \cite{gori2005gnn} provided a way to assemble neural modules over graph-like structures which was later applied to many different domains including ranking webpages, matching subgraphs, and recognising mutagenic compounds. In this first iteration each node $n$ would have a state $x_{n}$, iteratively updated through the application of a parametric function $f_{w}$, which receives as input both the node's label $l_{n}$ as well as the state and label from the nodes in its neighbourhood $N(n)$, such that the update from a node in iteration $t+1$ is calculated as in Equation~\ref{eq:gnn}. The way this model was intended to work, as the authors first envisioned, is that this updating would take place until the node's states reach a fixed point, after which they would be used for the  solution.

\begin{equation}\label{eq:gnn}
x_{n}^{t+1} = f_{w}(l_{n},x_{N(n)}^{t},l_{N(n)})
\end{equation}

This model was then later generalised to provide support for different types of entities and relations \cite{scarselli2009gnn}, which makes it general enough to be seen as a the first full realisation of GNN's potential. There have been two main viewpoints used to describe GNNs in the literature recently: that of message-passing neural networks and that of convolutions on graphs. In this paper we focus on the graph convolutional viewpoint, more specifically on the one presented in \cite{kipf2017gcn}. We do not specify any equations for the MPNN viewpoint as this is trivially transferable from what is presented here.

The idea of allowing convolutions over relational data stems from the concept that discrete spatial convolutions, widely used in the context of convolutions over images, are themselves a subset of convolutions in an arbitrary relational space, such as a graph or hypergraph, only being restricted to the subset of grid-like graphs \cite{wu2019comprehensive}. This idea gave rise to many different formalisations and models that applied convolutions over relational data, which are classified \cite{wu2019comprehensive} into spectral-based and spatial-based. Here, we refer to spectral-based approaches as graph convolutional networks (GCNs). The model proposed in \cite{kipf2017gcn} defines approximate spectral graph convolutions and apply them to build a layered model to allow the stacking of multiple convolutions, as defined in Equation~\ref{eq:gcn} below, where $\Tilde{D_{(i,i)}} = \sum_{j}\Tilde{A{(i,j)}}$ is a normalisation component that is used to divide the incoming embedding for each vertex in the graph by its degree, $\Tilde{A} = A + I_{N}$ is the adjacency matrix (with added self-connections to allow the node to keep its own information) $\sigma$ is any activation function, and $W^{l}$ is the weight kernel for layer $l$.

\begin{equation}\label{eq:gcn}
H^{l+1} = \sigma( \Tilde{D}^{-\frac{1}{2}}\Tilde{A}\Tilde{D}^{-\frac{1}{2}}H^{l}W^{l} )
\end{equation}

Such model is a simple, yet elegant, formalisation of the notion underlying  graph convolutions. It allows one to stack multiple layers, which has been argued as one of the ways to improve model complexity and performance in deep learning \cite{he2016resnet}, but it has been shown that stacking more layers can decrease performance on GCNs \cite{li2018gcndeeper}, which was one of the main motivators for applying continuous residual modules in this paper.

\section{Designing Residual Graph Kernels}\label{sec:res}

The main idea behind Residual Networks is that the network is made to learn a residual function instead of a whole transformation \cite{he2016resnet,greff2017highwayresidual}. In this way, a module which would work as in Equation~\ref{eq:idea_nonres} is then transformed as in Equation~\ref{eq:idea_res}, where $H^{l}$ denotes the input tensor and $W^{l}$ the function parameters at layer $l$ in the neural network.

\begin{equation}\label{eq:idea_nonres}
H^{l+1} = f(H^{l}, W^{l})
\end{equation}

\begin{equation}\label{eq:idea_res}
H^{l+1} = f(H^{l}, W^{l}) + H^{l}
\end{equation}

While this idea can seem too simplistic to bring any benefits, it has since been proven to improve performance in many different meta-models \cite{he2016resnet,kim2017reslstm,wang2016resrnn}, and has been used to allow one to build CNNs with more layers than traditionally. As stated in Section~\ref{sec:gnn}, we wanted to be able to benefit similarly from residual connections with graph data. One way to visualise how this change can help is that the function learned by the model is as an Euler discretisation of a continuous transformation \cite{chen2018odenet,haber2017continuous,lu2018continuous,ruthotto2018continuous}. So instead of learning a full transformation of the input, it learns to map the derivative of the input, as shown rearranged in Equation~\ref{eq:idea_res_derivative} below\footnote{We use function notation for the continuous residual modules and their derivations to make the derivative more explicit}.

\begin{equation}\label{eq:idea_res_derivative}
f(H(l), W(l), l) = \frac{H(l+1) - H(l)}{(l+1)-l} \approx \frac{\delta H(l)}{\delta l}
\end{equation}

In this section, we will focus on the intuition behind discrete and continuous residual layers on Graph Convolutional Networks.

\subsection{Residual Modules on Canonical CNNs}

One of the first successes of this technique has been the use of such a kernel in the context of convolutional neural networks applied over image data \cite{he2016resnet}. It has been argued that this technique allows the networks to increase in depth while maintaining or reducing the parameter space learned, since each module has to learn only the transformation to be applied to the input instead of both the transformation and the application of such transformation. In the same vein, the residual connections create shortcuts for the gradients to pass through, reducing the exploding/vanishing gradient problem for larger networks. All this helps accelerating convergence and improves the overall performance of the model, while still allowing one to perform more complex operations on data.

Many different modules have been proposed and tested with this technique. One caveat, however, is that both the input and output of a residual module must either have the same dimensionality, or be expanded/contracted with arbitrary data to match the dimensionality of each other.

\subsection{Discrete Residual Modules on GCNs}

One of the easiest GNN models from which we can extend the idea of a Residual block is the one based on graph convolutions. Here, we focus on the canonical model proposed in \cite{kipf2017gcn} and explained in Section~\ref{sec:gnn}. For our experiments we use a slightly modified version, which does not perform symmetric normalisation, computing $\Tilde{D}^{-1}\Tilde{A}$ instead of $\Tilde{D}^{-\frac{1}{2}}\Tilde{A}\Tilde{D}^{-\frac{1}{2}}$, which will be used as the baseline for this technique in Section~\ref{sec:exp}, but we keep the original presented in its original form \cite{kipf2017gcn}.

We argue that the GCN model is the easiest to re-frame into a Residual block since it is both based on the notion of convolution and provides as output a tensor with the same number of nodes as the input values -- i.e. does not reduce the number of elements to be processed in the next feature map's shape. The transformation of such a module into a residual one can be achieved by simply engineering it to contain the residual input, such as in Equation~\ref{eq:rgcn}.

\begin{equation}\label{eq:rgcn}
H(l+1) = H(l) + \sigma( \Tilde{D}^{-\frac{1}{2}}\Tilde{A}\Tilde{D}^{-\frac{1}{2}}H(l)W(l) )
\end{equation}

\subsection{Continuous Residual Modules for Graphs}

Recently, Chen et. al. \cite{chen2018odenet} proposed a model which approximates a continuous-time (or continuous-layer) derivative function which can be efficiently integrated through parallel ODE solvers. These models are generated through taking the approximation presented in Equation~\ref{eq:idea_res_derivative} and using ODE solvers to integrate them as needed, effectively learning a Lipschitz-continuous function that can be efficiently evaluated at specified points for producing results regarding to those points.

In terms of residual layers, the learned derivative function can be seen as producing a function that is continuous in the layer-space -- that is, they produce a continuous equivalent of the non-residual layer. Furthermore, they provide a way to generate a continuous function on the layers themselves, tying nearby layer weights to each other while allowing for different transformations to be applied in each of them. If one sees these as recurrent functions, they can also be seen as producing recurrent networks that work in continuous spaces, instead of needing to use discretely sampled application of the recurrent network one can simply evaluate it at the required times.

This idea can easily be applied to graph convolutional layers by producing a continuous equivalent of Equation~\ref{eq:rgcn}, as shown in Equation~\ref{eq:odegcn}. With this, one arranges the graph convolutional modules in different graph configurations and solve the differential equations given this structural format.

\begin{equation}\label{eq:odegcn}
\frac{\delta H(l+1)}{\delta l} = \sigma( \Tilde{D}^{-\frac{1}{2}}\Tilde{A}\Tilde{D}^{-\frac{1}{2}}H(l)W(l) )
\end{equation}

Since, in the problem we consider, the graph structure is independent of the layer-space, we can set this part of the function ($\Tilde{D}^{-\frac{1}{2}}\Tilde{A}\Tilde{D}^{-\frac{1}{2}}$) as fixed on every batch and through each pass in the ODE solvers. In this learned function continues to be free of the graph structure for its application, using it only as a structure through which propagate information that is accumulated in the repeated neural modules for each node. A simple way to visualise this is to imagine a mass-spring system expressed as a graph: The learned function will then learn the dynamics of the mass-spring system for many different configurations, leading to it being useful in differently sized and arranged systems. This mass-spring intuition is the same used to explain Interaction Networks \cite{battaglia2016interaction} and the Graph Network formalisation \cite{battaglia2018gn}.

\subsection{Multiple layers in Constant-Memory}\label{sec:res:sub:constant}

Chen et. al. \cite{chen2018odenet} argue that the technique of allowing continuous-layer\footnote{In the remainder of this paper we refer only to layers and layer-space for the CNN viewpoint, but one could re-interpret this as time in a RNN} residual layers makes it possible to build a many-layered model in constant space instead of quadratic. That is, instead of stacking $k$ layers with $d \times d$ dimensions for each kernel, one could build a single residual layer with $(d+1) \times (d+1)$ dimensions, with the extra dimension being the layer component of the model. The intuition behind this is that the model parameter space will become dependant on the layer-space, with this it can behave differently when evaluated on a point in the layer-space. This can be visualised in the difference between Equations \ref{eq:nondependantweight} and \ref{eq:dependantweight}. In Equation~\ref{eq:nondependantweight} the learned kernel $W$ has a dimension $d \times d$, and we would need to stack $k$ of such layers to produce $k$ different transformations, whereas in Equation~\ref{eq:dependantweight} the kernel $W'$ has $(d+1) \times (d+1)$ dimensions. These continuous pseudo-layers can then be evaluated in as many points as warranted in the ODE solver, effectively allowing a dynamic number of layers to be computed instead of a singular discrete composition.

\begin{equation}\label{eq:nondependantweight}
f(H(l)) = H(l)W(l)
\end{equation}

\begin{equation}\label{eq:dependantweight}
f(H(l),l) = \operatorname{concat}(H(l),l)W'(l)
\end{equation}

This technique, however, enforces that those pseudo-layers behave similarly for close points in the layer-space, effectively making them continuous. This constraint both forces the learned transformations to be closely related in the layer space as well as makes it so that the composition of these various layers is relatively well-behaved. Thus, we can expand the number of evaluated layers dynamically by choosing more points to integrate in. And even if we fix the start and end-points for the integration over the layer-space, the learned network can be integrated in many points between these to provide an answer with the accuracy required from the ODE solver.

Whenever we apply continuous residual layers in this work, we make use of this technique to allow the ODE solver to change the layer transformation slightly between each point in the layer-space. Thus, one could consider that the ODE-solved models we present in the results have more layers than reported, for this we argue that this difference is at most linear when the residual layers consists of only a single residual GCN application, since the GCN layers themselves are single-layered and the additional layer-space value provided as input can only interfere in this linear application through its weights in the kernel matrix multiplication. We also believe a similar technique could, in theory, be applied without the use of an ODE-solver to integrate through the layers, but one would lose the benefits of the ODE solver being able to define by itself which points need to be evaluated.

\section{Experiments on semi-supervised classification in citation networks}\label{sec:exp}

In this section we evaluate the transformations discussed in Section~\ref{sec:res} to small adaptations of GCN neural modules  described in \cite{kipf2017gcn}.
The task of interest is semi-supervised classification in citation networks, where nodes are scientific papers and edges are citation links, and only a small fraction of the nodes is labelled.
The experiments are as in \cite{kipf2017gcn}, with the same train/test/evaluation split (inherited from \cite{yang2016semisupsplits}) in Cora, Citeseer and Pubmed citation networks.
They have 6, 7 and 3 classes, respectively.

To capture the difference in performance and stability due to applying residual blocks to GNNs, we adapted the Pytorch code of the original GCN paper \footnote{See \url{https://github.com/tkipf/pygcn} for the model and \url{https://github.com/tkipf/gcn} for the datasets and test/train/evaluation splits.} \cite{kipf2017gcn}, changing the initialisation, degree normalisation, and removing dropout on the input features in our GCN kernels.
Moreover, our GCN model follows Equation~\ref{eq:gcnassymetric} for the layer-wise propagation rule instead of the original one. Similarly, the discrete residual module follows Equation~\ref{eq:rgcnassymetric}, and the continuous one approximates Equation~\ref{eq:odeassymetric}.

\begin{equation}\label{eq:gcnassymetric}
H^{l+1} = \sigma( \Tilde{D}^{-1}\Tilde{A}H^{l}W^{l} )
\end{equation}

\begin{equation}\label{eq:rgcnassymetric}
H^{l+1} = \sigma( \Tilde{D}^{-1}\Tilde{A}H^{l}W^{l} )
\end{equation}

\begin{equation}\label{eq:odeassymetric}
\frac{\delta H(l)}{\delta l} = \sigma( \Tilde{D}^{-1}\Tilde{A}H(l)W(l) )
\end{equation}

\subsection{Experimental Setup}

All experiments were developed in Python v3.6.7, and the libraries and frameworks used were Pytorch v1.0.1.post2, Scipy v1.2.1, Numpy v1.16.3, Matplotlib v3.0.3 and NetworkX v2.3. 
The experiments where time was reported were executed in a computer with a Intel\textsuperscript{\textregistered} Core\textsuperscript{\texttrademark} i7-8700 @ $3.2$ GHz, with 32GB RAM, a NVIDIA\textsuperscript{\textregistered} Quadro\textsuperscript{\texttrademark} P6000 with Cuda\textsuperscript{\texttrademark} 10.1. 
Some experiments were also executed in a computer with a Intel\textsuperscript{\textregistered} Core\textsuperscript{\texttrademark} i7-7700 @ 2.8 GHz, with 32GB RAM, a NVIDIA\textsuperscript{\textregistered} GeForce\textsuperscript{\texttrademark} 1070 with Cuda\textsuperscript{\texttrademark} 9.2.

\subsection{Three-layered models}\label{sec:exp:sub:gcn}

In these experiments, we built neural networks with three graph convolutional layers whose feature dimensionalities were $(h,h,c)$, with $h$ being an hyperparameter of the model and $c$ the number of classes in the dataset.

We initially evaluate seven models, and run subsequent experiments for the best three. 
We also include a closer replication of \cite{kipf2017gcn} in Section~\ref{sec:exp:sub:gcnpaper}, with discussion on the main differences between this and the original paper. 
The seven initially tested models use either dropout \cite{hinton2012dropout}, group normalisation \cite{wu2018groupnorm} (or both), and L2 normalisation of the parameters, as follows:

\begin{description}
    \item[GCN-3] A model with the GCN layer as made available by \cite{kipf2017gcn}, with dropout applied between each pair of layers.
    \item[GCN-norm-3] Equivalent to GCN-3, but with dropout applied between the first and second layer and group normalisation applied between the second and the third.
    \item[RES-3] A model with a residual GCN kernel as defined in Equation~\ref{eq:rgcnassymetric} instead of a normal GCN on the second layer, with dropout applied between each pair of layers.
    \item[RES-norm-3] Equivalent to RES-3, but with dropout applied between the first and second layer and group normalisation applied between the second and the third.
    \item[RES-fullnorm-3] Equivalent to RES-3, with group normalisation applied between each pair of layers.
    \item[ODE-norm-3] A model with a continuous residual module as defined in Equation~\ref{eq:odeassymetric} instead of a normal GCN on the second layer, dropout before the ODE-solved layer and group normalisation as part of the ODE-solved layer, applied to its input. The ODE-solved layer use the technique described in Section~\ref{sec:res:sub:constant}
    to allow the learned continuous transformation to be dependant on the time parameter evaluations.
    \item[ODE-fullnorm-3] Contains the same continuous residual module of ODE-norm-3 on the second layer, but group normalisation both before and as part of the ODE-solved layer, applied to its input. 
    The ODE-solved uses the same technique of ODE-norm-3.
\end{description}

Having constructed the networks above, we ran the experiments of \cite{kipf2017gcn} for semi-supervised classification in the Cora, Citeseer and Pubmed citation networks, using the same train-validation-test splits, over 2500 runs in the discrete models and 250 in the continuous ones, averaging the results to minimise the influence of random parameter initialisation. 
All models were trained with $h = 16$, which was kept from the original code as the default, since \cite{velickovic2018gat} showed that increasing the number of features did not seem to improve performance on the GCN, a learning rate of $0.01$, $50\%$ dropout and L2 normalisation on the weights, scaled by $5\times10^{-4}$. 
All learned kernels weights and biases are initialised with the uniform distribution $\mathcal{U}(-\sqrt{k}, \sqrt{k})$, where $k = \frac{1}{\text{out\_features}}$.

Table~\ref{tab:gcn3-vs-rgcn3-full} shows the average, standard deviation, best (max) and worst (min) values over all the runs for accuracy as well as average loss and runtime. 
There, one can see that the residual models have a consistently better performance, as well as less variance.
The residual modules heavily benefited from group normalisation, however they were slowed by this addition.
The continuous GCN model achieved the best average accuracy in Cora and Pubmed, and was close to the best in Citeseer. However, it was much slower, partly due to the group normalisation inside the integrated function. 
We tried to train an ODE model with dropout in the integrated function or without any normalisation, but it failed to converge in the first case and severely overfitted in the second. 
Even if we consider only the best over all runs, RES-norm-3 performed better than any GCN-3 variant, and ODE-norm-3 was less sensitive to weight initialisation, by showing a consistently lower standard deviation.

\begin{figure}[htpb]
    \centering
    \includegraphics[width=0.9\linewidth]{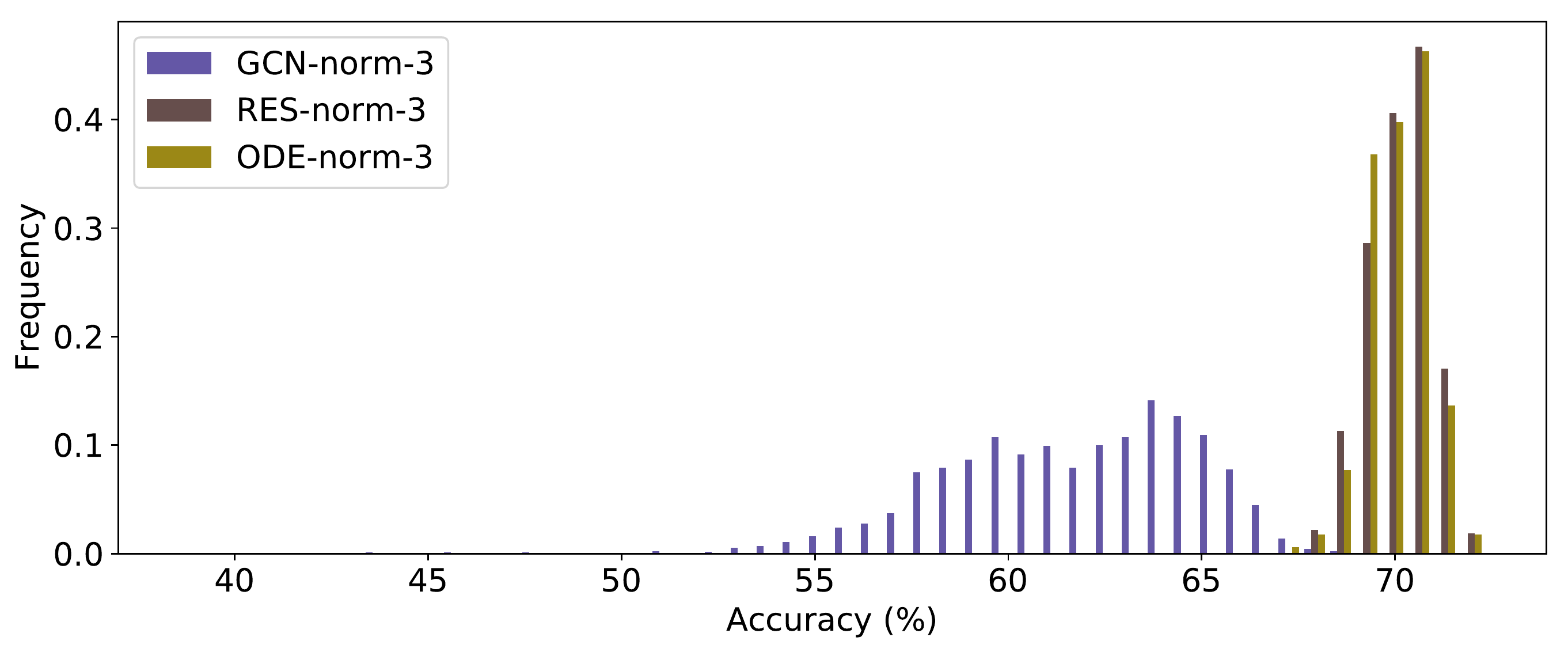}
    \caption{50-bin histogram of the accuracies, comparing the models on the \bemph{Citeseer} dataset.}
    \label{fig:gcn-vs-rgcn-hist-citeseer}
\end{figure}

\begin{figure}[htpb]
    \centering
    \includegraphics[width=0.9\linewidth]{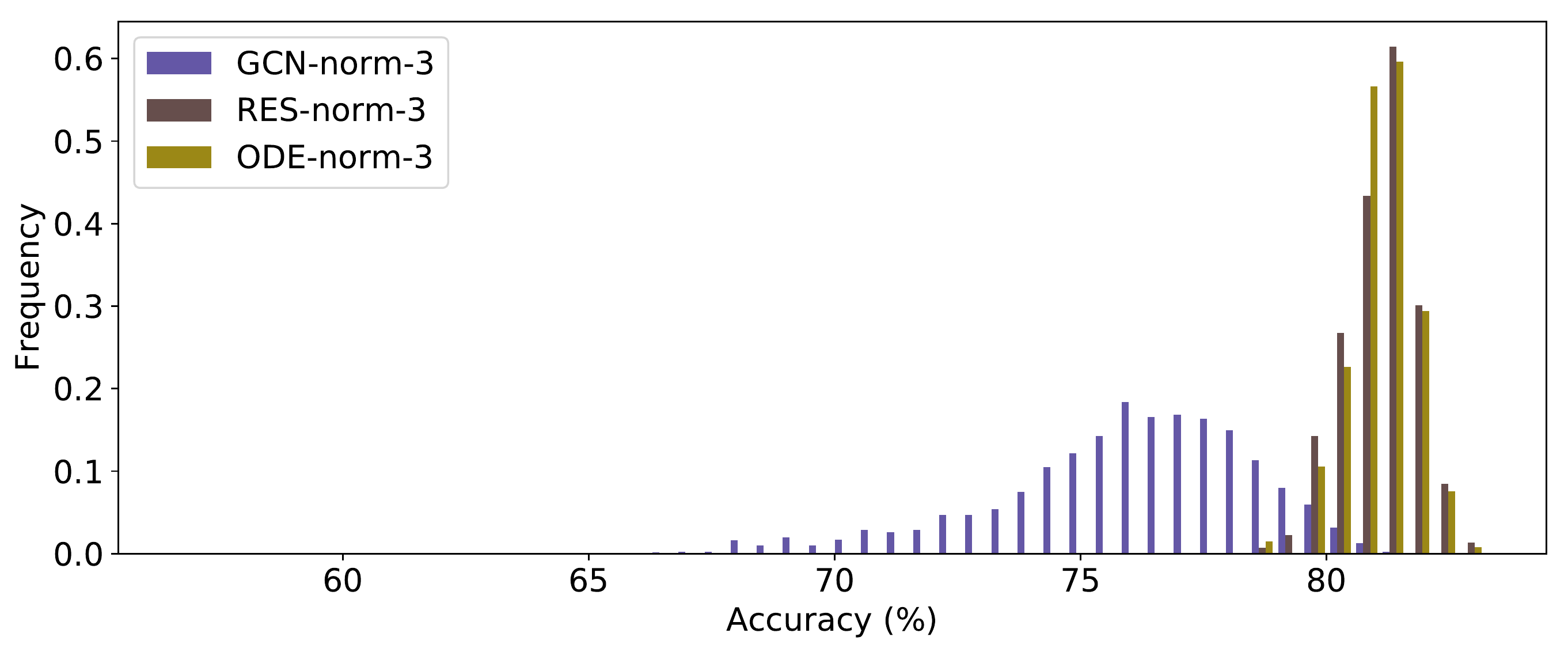}
    \caption{50-bin histogram of the accuracies, comparing the models on the \bemph{Cora} dataset.}
    \label{fig:gcn-vs-rgcn-hist-cora}
\end{figure}

\begin{figure}[htpb]
    \centering
    \includegraphics[width=0.9\linewidth]{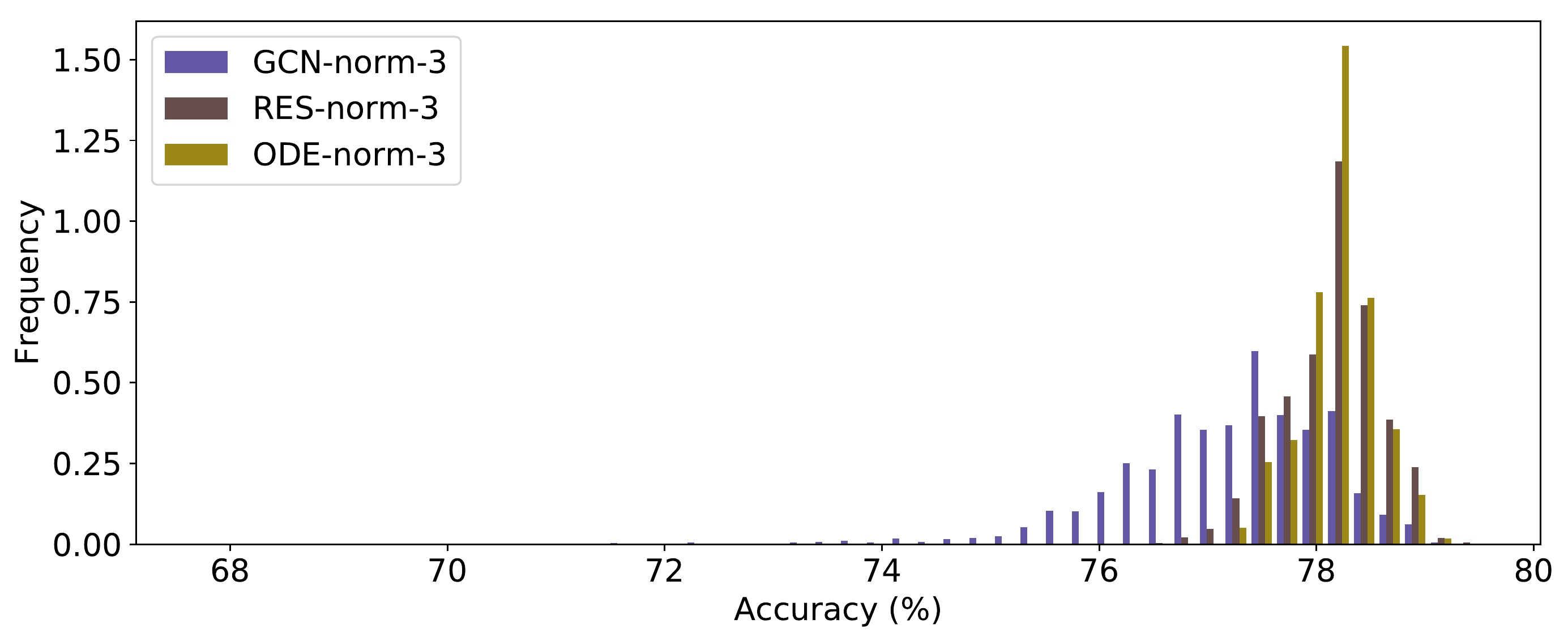}
    \caption{50-bin histogram of the accuracies, comparing the models on the \bemph{Pubmed} dataset.}
    \label{fig:gcn-vs-rgcn-hist-pubmed}
\end{figure}

To further validate these results, we ran statistical tests on the accuracies to see whether the differences between non-residual and residual layers were statistically significant, the p-values for the Mann-Whitney U-test and Kruskal-Wallis H-test were both lower than $10^{-10}$ in the \bemph{Pubmed} dataset, and even lower in the other two, when comparing a residual (RES, RES-norm and ODE, ODE-norm) module with a non-residual module (GCN, GCN-norm). 
The performance of the discrete and continuous residual modules was statistically similar, with p-values higher than 5\% for all datasets. Figures \ref{fig:gcn-vs-rgcn-hist-citeseer}, \ref{fig:gcn-vs-rgcn-hist-cora} and \ref{fig:gcn-vs-rgcn-hist-pubmed} show the histograms of the accuracies over these runs for each ``norm'' model, and can help in visualising that the residual ones are significantly better in average.

\begin{table}
    \scriptsize
    \centering
    \setlength{\tabcolsep}{4pt}
    \begin{tabular}{ccccccc}
        \toprule
        \multirow{2}{*}{Model} & \multicolumn{4}{c}{Acc (\%)} & Loss & Time (s) \\
         & Avg & Std & Min & Max & Avg & Avg \\
        \midrule
        \multicolumn{7}{c}{\bemph{Citeseer}} \\
        \midrule
        Presented in \cite{kipf2017gcn} & \textit{70.30} & - & - & - & - & 7 \\
        GCN-3 & 61.70 & 3.32 & 37.20 & 68.80 & 1.3344 & \textbf{1.4325} \\
		GCN-norm-3 & 61.66 & 3.29 & 38.60 & 68.70 & 1.3356 & 1.4399 \\
		RES-3 & 65.87 & 1.46 & 58.10 & 70.10 & 1.1069 & 1.4480 \\
		RES-norm-3 & \textbf{70.08} & 0.79 & 67.40 & \textbf{72.30} & \textbf{1.0132} & 2.2851 \\
		RES-fullnorm & 16.17 & 4.99 & 7.70 & 23.10 & 1.7918 & 3.1579 \\
		ODE-norm-3 & 70.04 & \textbf{0.72} & \textbf{67.50} & 71.80 & 1.0163 & 69.7444 \\
		ODE-fullnorm-3 & 18.28 & 2.59 & 16.00 & 23.10 & 1.7918 & 61.0533 \\
        \midrule
        \multicolumn{7}{c}{\bemph{Cora}} \\
        \midrule
        Presented in \cite{kipf2017gcn} & \textit{81.50} & - & - & - & - & 4 \\
		GCN-3 & 76.01 & 2.59 & 56.70 & 81.50 & 0.8554 & \textbf{1.3841} \\
		GCN-norm-3 & 75.95 & 2.68 & 56.70 & 81.70 & 0.8554 & 1.3944 \\
		RES-3 & 78.98 & 1.32 & 70.60 & 82.20 & \textbf{0.7114} & 1.3888 \\
		RES-norm-3 & 81.06 & 0.72 & \textbf{78.70} & \textbf{83.20} & 0.7275 & 2.0943 \\
		RES-fullnorm & 15.07 & 8.87 & 6.40 & 31.90 & 1.9459 & 2.7927 \\
		ODE-norm-3 & \textbf{81.08} & \textbf{0.67} & 78.60 & 82.70 & 0.7333 & 62.2312 \\
		ODE-fullnorm-3 & 14.09 & 6.49 & 6.40 & 31.90 & 1.9458 & 54.8411 \\
        \midrule
        \multicolumn{7}{c}{\bemph{Pubmed}} \\
        \midrule
        Presented in \cite{kipf2017gcn} & \textit{79.00} & - & - & - & - & 38 \\
		GCN-3 & 77.19 & 1.01 & 68.10 & 79.30 & 0.7378 & \textbf{5.6163} \\
		GCN-norm-3 & 77.19 & 0.99 & 67.70 & 79.30 & 0.7375 & 5.6146 \\
		RES-3 & 77.45 & 0.77 & 74.20 & 79.20 & 0.7081 & 5.6194 \\
		RES-norm-3 & 78.13 & 0.44 & 76.10 & \textbf{79.50} & \textbf{0.5602} & 10.5187 \\
		RES-fullnorm & 32.82 & 11.11 & 18.00 & 41.30 & 1.0986 & 15.4539 \\
		ODE-norm-3 & \textbf{78.18} & \textbf{0.34} & \textbf{77.20} & 79.20 & \textbf{0.5602} & 346.6378 \\
		ODE-fullnorm-3 & 36.40 & 9.20 & 18.00 & 41.30 & 1.0986 & 289.2936 \\
        \bottomrule
    \end{tabular}
    \caption{Comparison of the performance in the reproduction of the experiments of \cite{kipf2017gcn}. The experiments were run 2500 times for the non-continuous models (those that don't start with ``ODE''), and 250 times for the continuous ones. We report the average, standard deviation, minimum and maximum accuracy of these runs to minimise the effect of the variables' random initialisation. The best model (except the original of \cite{kipf2017gcn}) is marked in \textbf{bold} for each metric. Runtime of the original paper is presented for completeness, as it was obtained in a different setup.}
    \label{tab:gcn3-vs-rgcn3-full}
\end{table}

\subsection{k-layered models}

For the second battery of tests, we present the results for the \bemph{Pubmed} dataset\footnote{We only present for the Pubmed since it is the best case for the baseline non-residual model, as per Figure~\ref{fig:gcn-vs-rgcn-hist-pubmed}.}. 
The ``K'' models work in the same way as the ``3'' models from Section~\ref{sec:exp:sub:gcn}, except that they have K layers instead of 3, with the normalisation between the first and second layer being the same, and on the other layers being the same as the normalisation between the second and third layer in the ``3'' model. All models had ReLU activations after every layer but the last, applied before the normalisation. On the last layer all models had a log softmax applied to each node's output. The residual modules with connections every two layers had a residual connection on the last layer if the number of layers is odd. The ODE-solved modules with residual connections every two layers used a time component as input to both layers, appended to every node's feature vector. All ODE models are solved using the adjoint method described in \cite{chen2018odenet}.

\begin{figure}[tpb]
    \centering
    \includegraphics[width=.95\linewidth]{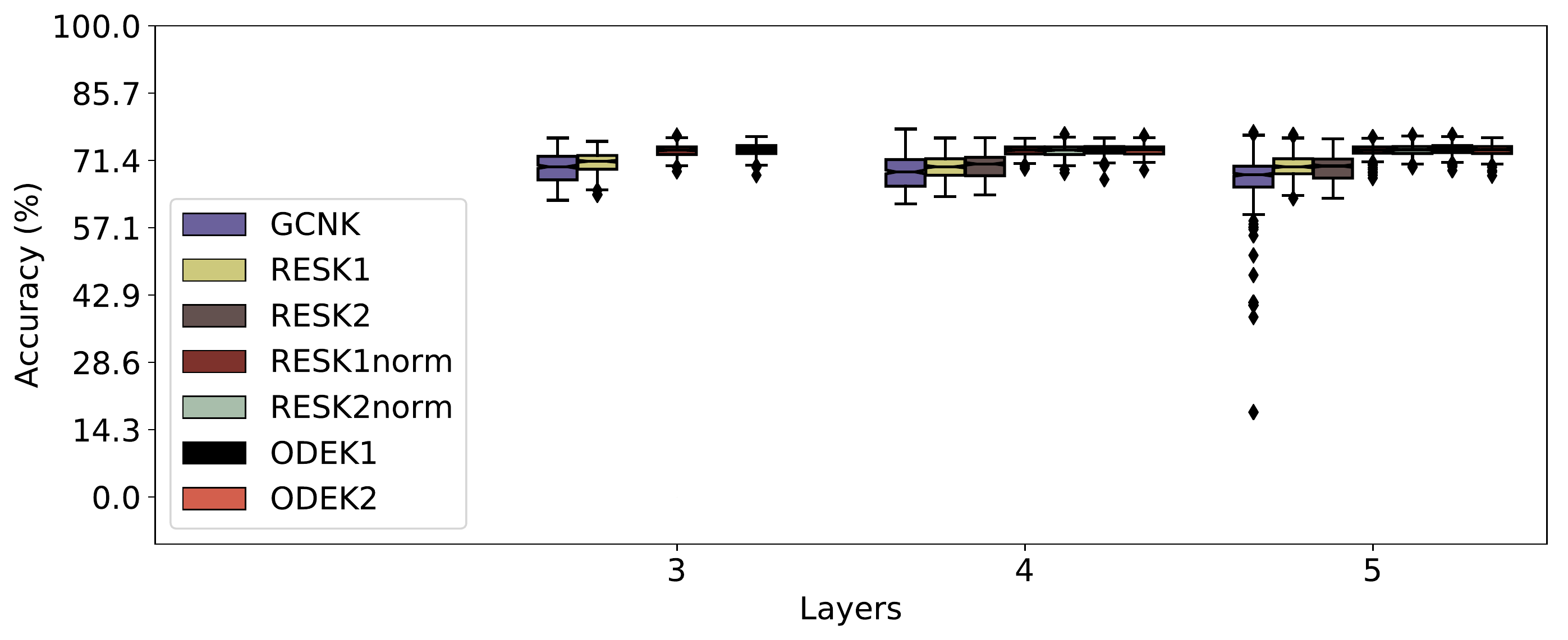}
    \caption{Average final test accuracy of the models which hit the early stopping criteria in the \bemph{Pubmed} dataset.}
    \label{fig:pubmed-c-acc}
\end{figure}

\begin{figure}[tpb]
    \centering
    \includegraphics[width=.95\linewidth]{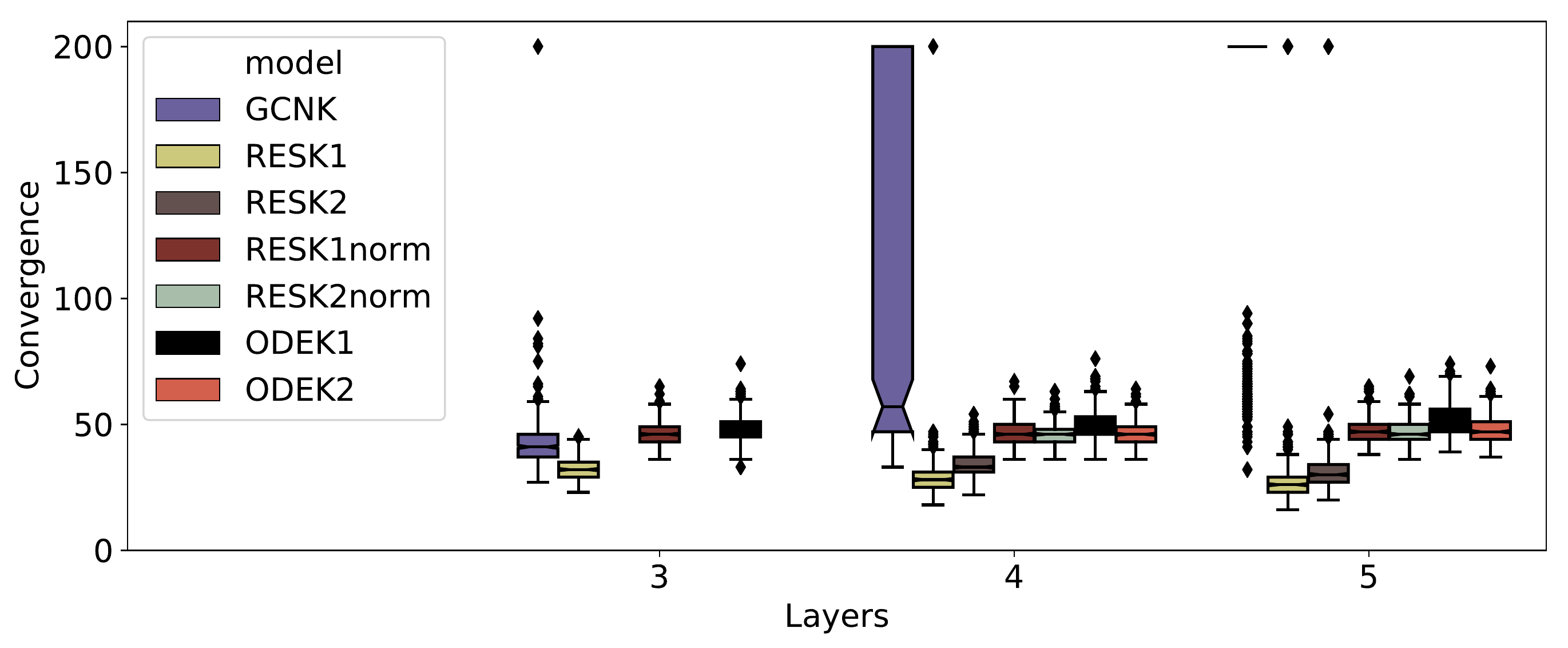}
    \caption{Average number of iterations that the models hit the early stopping criteria in the \bemph{Pubmed} dataset, stopping at a maximum of 200 epochs.}
    \label{fig:pubmed-c-iter}
\end{figure}

\begin{figure}[tpb]
    \centering
    \includegraphics[width=.95\linewidth]{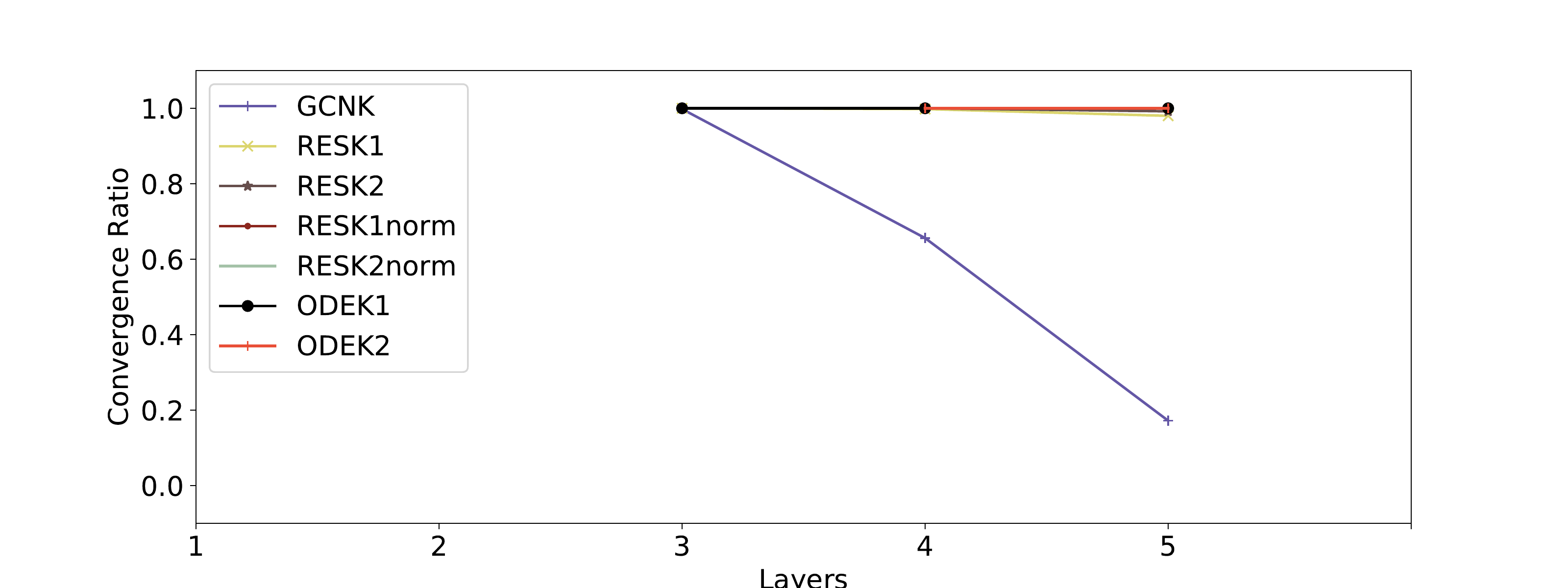}
    \caption{Ratio of models that hit the early stopping criteria in the \bemph{Pubmed} dataset.}
    \label{fig:pubmed-c-ratio}
\end{figure}

We trained the models for a maximum of 200 epochs, stopping earlier if the validation accuracy of the model was higher than $90\%$ of the lowest accuracy of the models in Table~\ref{tab:gcn3-vs-rgcn3-full} and the validation loss was lower than $110\%$ of the highest test loss obtained during the same test. As one can see in Figures \ref{fig:pubmed-c-acc}, \ref{fig:pubmed-c-iter}, \ref{fig:pubmed-c-ratio}, in these tests the many-layered non-residual models often failed to converge before the defined maximum number of epochs and had a worse performance when compared to the residual models. Figure~\ref{fig:pubmed-c-iter}, in particular, shows that the residual models converge to a better validation accuracy (and thus hit the early stopping criteria) at less than half the maximum number of iterations proposed in~\cite{kipf2017gcn}, while also some also show to be more immune or even benefit from more layers to converge faster. 
We also ran this experiment for more training iterations and deeper networks: the non-residual models were all prone to overfitting while the residual models were more or less immune to it, achieving good test accuracy even the earlier stopping criteria was not met.

The performance degradation caused by stacking more layers may be also partially explained due to the fact that the early stopping threshold included not only an accuracy threshold but also a loss one, which may have caused some of the models to overfit the training data. The reason for using a loss threshold along the accuracy on the validation set is that we wanted our model to be confident enough about its predictions and not only accurate. We also ran this experiment for a larger number of layers and the results seemed stable throughout, we chose to present here only from layers 3 through 5 since in this range the performance degradation of non-residual GCNs is already visible.

\subsection{Details on replicating the GCN experiment}
\label{sec:exp:sub:gcnpaper}

Note that all the experiments we've done here, with three-layered networks, perform slightly worse than a two-layered network (the original of \cite{kipf2017gcn}) in most datasets. The original paper already shows that two seems to be the optimal number of layers for this dataset, and in the original paper they used a different kernel initialization method. The main point of our experiments was to show the immunity of the residual networks to the number of layers and initial parameter intialisation. We trained a two-layered discrete residual network, taking only a slice of the output of the layer as the final features\footnote{This was done so that the layer has the same number of in and out features}, this model performed similarly to the non-residual module, and achieved performance near to the one presented in the original paper.

\begin{figure}[tpb]
    \centering
    \includegraphics[width=.95\linewidth]{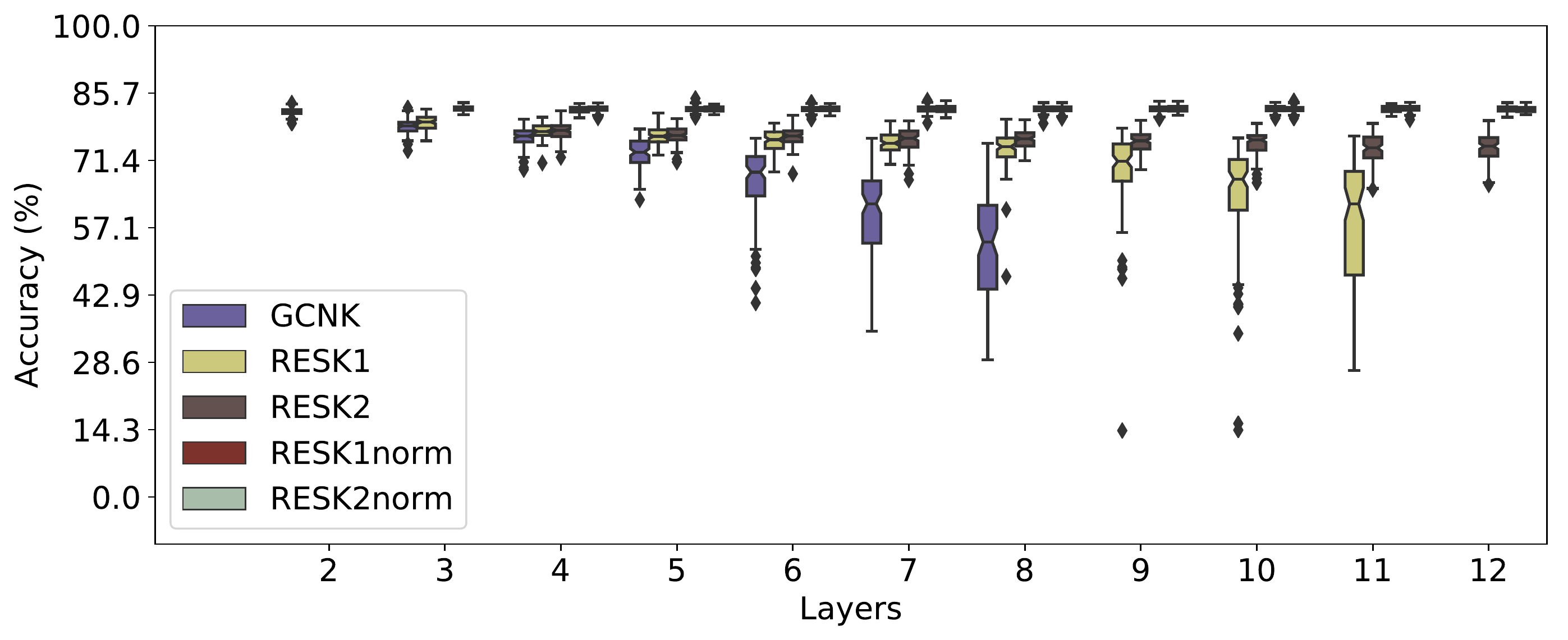}
    \caption{Average final test accuracy of the models which hit the early stopping criteria in the \bemph{Cora} dataset by following the paper more closely.}
    \label{fig:paper-cora-c-acc}
\end{figure}

Another difference between what we present here and the results originally published (one of the parts where the code we used was ``subtly different'' from the one which produced the published results) is that the original paper used a different kernel initialisation (the Xavier/Glorot initialisation described in \cite{glorot2010init}). We tested the models with the Glorot initialisation but the models still seemed to slightly underperform the results shown in the original paper.

\begin{table}
    \centering
    \footnotesize
    \setlength{\tabcolsep}{4pt}
    \begin{tabular}{cccccc}
        \toprule
        \multirow{2}{*}{Model} & \multicolumn{4}{c}{Acc (\%)} & Loss \\
         & Avg & Std & Min & Max & Avg \\
        \midrule
        \multicolumn{6}{c}{\bemph{Citeseer}} \\
        \midrule
        Presented in \cite{kipf2017gcn} & \textit{70.30} & - & - & - & - \\
		GCN-3 & 65.71 & 2.04 & 55.60 & 69.10 & 1.1202 \\
		GCN-norm-3 & 65.49 & 1.98 & 56.50 & 69.30 & 1.1306 \\
		RES-3 & 66.78 & 1.39 & 63.10 & 69.80 & 1.0776 \\
		RES-norm-3 & 70.75 & 0.85 & 68.50 & 73.00 & 1.0433 \\
		ODE-norm-3 & 69.51 & 1.09 & 67.30 & 72.10 & 1.0616 \\
        \midrule
        \multicolumn{6}{c}{\bemph{Cora}} \\
        \midrule
        Presented in \cite{kipf2017gcn} & \textit{81.50} & - & - & - & - \\
		GCN-3 & 79.41 & 1.52 & 75.80 & 82.80 & 0.6776 \\
		GCN-norm-3 & 79.59 & 1.46 & 75.70 & 82.20 & 0.6748 \\
		RES-3 & 80.33 & 1.21 & 77.90 & 82.80 & 0.6469 \\
		RES-norm-3 & 81.87 & 0.70 & 80.10 & 83.50 & 0.7710 \\
		ODE-norm-3 & 81.52 & 0.75 & 79.20 & 83.10 & 0.7841 \\
        \midrule
        \multicolumn{6}{c}{\bemph{Pubmed}} \\
        \midrule
        Presented in \cite{kipf2017gcn} & \textit{79.00} & - & - & - & - \\
		GCN-3 & 77.49 & 0.78 & 75.20 & 79.00 & 0.7063 \\
		GCN-norm-3 & 77.41 & 0.88 & 75.30 & 79.00 & 0.7121 \\
		RES-3 & 77.59 & 0.87 & 75.30 & 79.20 & 0.6924 \\
		RES-norm-3 & 79.11 & 0.60 & 77.20 & 80.10 & 0.5679 \\
		ODE-norm-3 & 78.50 & 0.47 & 77.20 & 79.80 & 0.5904 \\
        \bottomrule
    \end{tabular}
    \caption{Comparison of the performance in the reproduction of the experiments done in \cite{kipf2017gcn}. The experiments were run 100 times for all models. The results shown here are the average, standard deviation, minimum and maximum accuracy of these runs, as well as the average loss.}
    \label{tab:gcn3-vs-rgcn3-paper}
\end{table}

\begin{figure}[tpb]
    \centering
    \includegraphics[width=.95\linewidth]{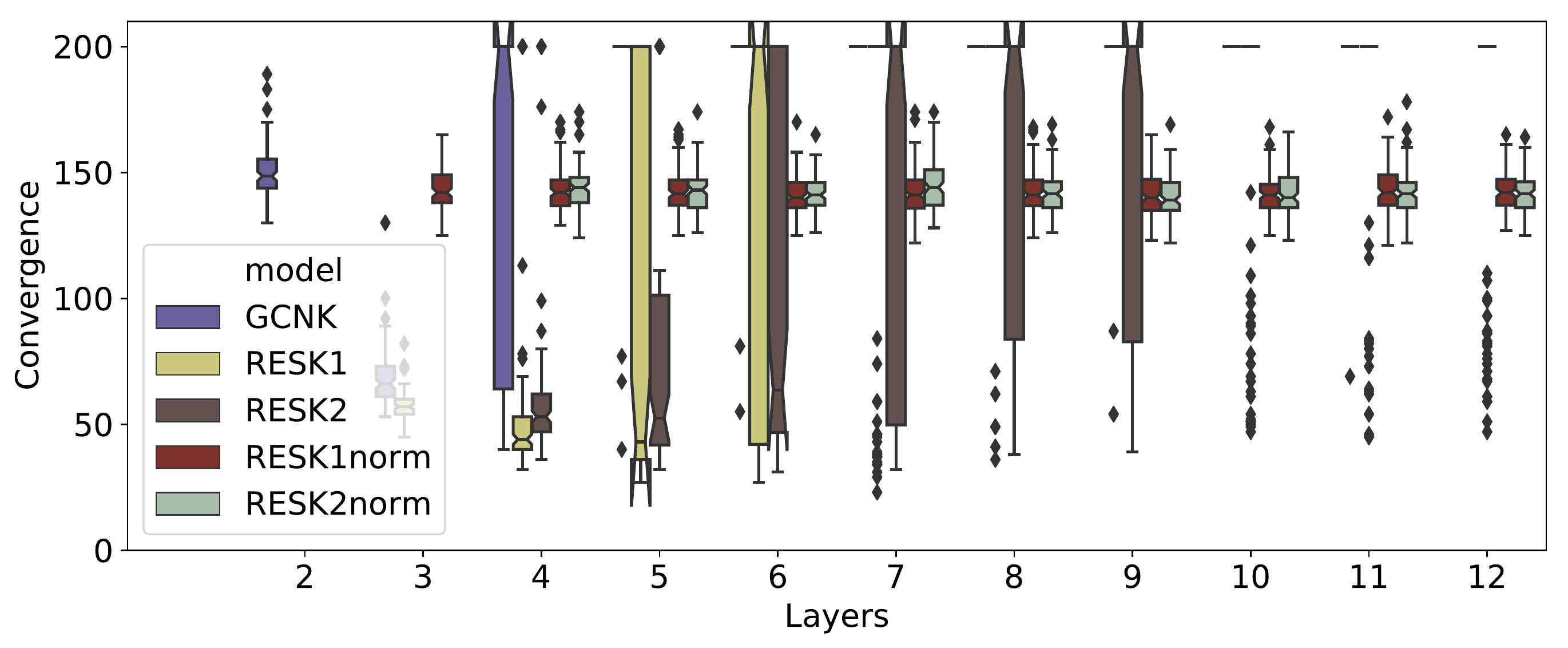}
    \caption{Average number of iterations that the models hit the early stopping criteria in the \bemph{Cora} dataset by following the paper more closely, stopping at a maximum of 200 epochs.}
    \label{fig:paper-cora-c-iter}
\end{figure}

\begin{figure}[tpb]
    \centering
    \includegraphics[width=.95\linewidth]{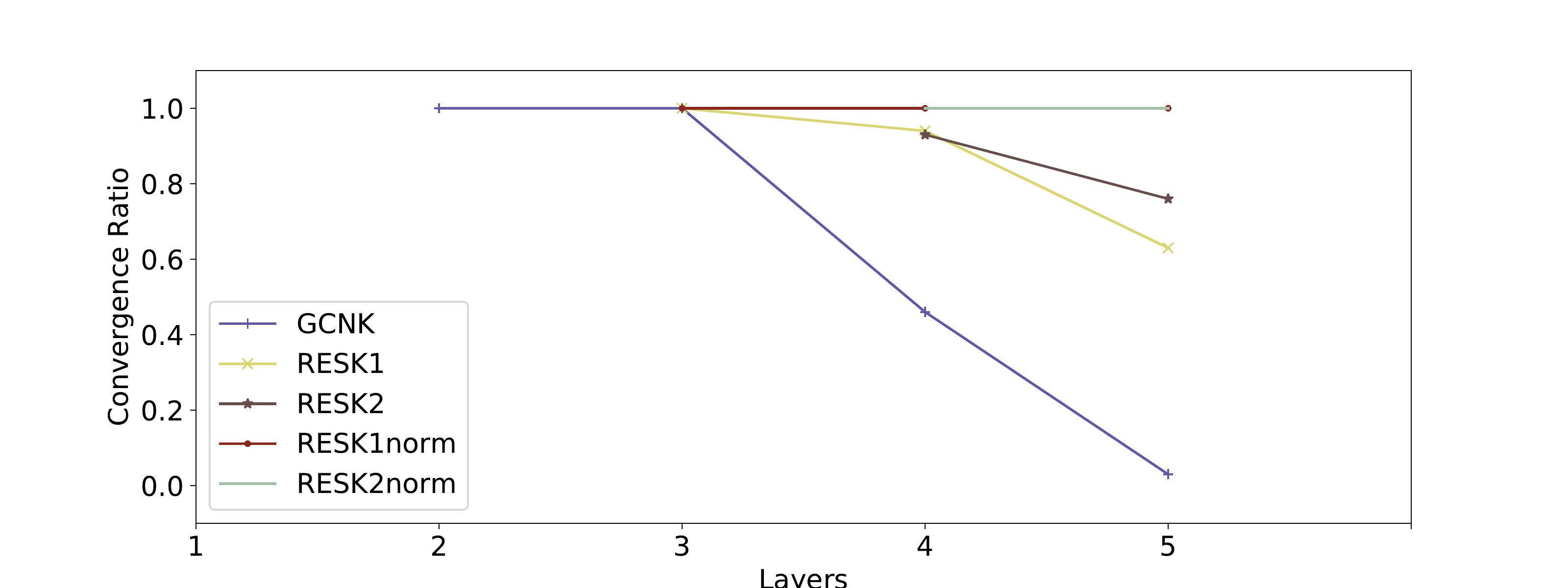}
    \caption{Ratio of models that hit the early stopping criteria in the \bemph{Cora} dataset by following the paper more closely.}
    \label{fig:paper-cora-c-ratio}
\end{figure}

The final trick to replicate the GCN experiment was to dropout in the input, whose results are in Table~\ref{tab:gcn3-vs-rgcn3-paper}. The two-layered GCN model achieved the same performance as in the original paper and the null hypothesis was rejected when comparing the GCN model to the ODE model, with the ODE model being slightly inferior than the GCN model. One can also look at Figures \ref{fig:paper-cora-c-acc}, \ref{fig:paper-cora-c-iter}, and \ref{fig:paper-cora-c-ratio} for results similar to the ones discussed in the other sections for the Cora dataset. 
These changes also allow more layers on the non-residual model before its performance degrades too much.

\section{Discussion}\label{sec:dis}

In this paper we provide, to the best of our knowledge, the first application of continuous-depth in a GNN. We engineer such a network by fixing the input graph before preforming the integral through an ODE solver. This creates a ODE system to be solved with the input graph's shape, without using the matrix as a input parameter to the ODE solver, which drastically reduces the memory usage. With this, the learned residual layer applies a continuous operation through the layer-space, which can behave better than using discrete transformations on the input.

Although the results we present here do not make such a strong case for the ODE-solved layers, we believe this to be mostly due to the application that the original GCN paper was applied to. The GCN model performed best with only two layers, which indicates that the datasets may not need, and may even be penalised by using, the information of a larger neighbourhood. 
Further experiments on problems which may require a more distant neighbourhood would surely showcase this model better, we nonetheless wanted to present our first results using the same dataset as the canonical GCN paper to provide an even footing, and to do this we utilised 3-layered models to allow the residual modules to learn features intrinsic to their feature space.

Furthermore, the stacking of ODE-solved layers as presented in the ``more layers'' experiments is not common, since the technique that allows multiple layers in constant memory is generally sufficient to produce an approximation of a many-layered network. Nonetheless, the stability seen throughout this application show that the benefit of residual learning still applies: allowing the network to keep its performance even when more layers are stacked.

The main advantage we believe that continuous residual layers can provide on graph-structured data would be to allow a more predictable behaviour on the learned functions, as was shown to be the case in other meta-models in \cite{chen2018odenet,gratwohl2018odenet}. This, we believe, would allow complex systems to be modelled as ordinary differential equations, which have a vast literature of theoretical analysis that could greatly benefit the Deep Learning community.

\section{Related Work}\label{sec:relwk}

The paper which provided the original GCN formalisation \cite{kipf2017gcn} already experimented with residual connections and showed that these allow training a GCN with more layers, but did not experiment with continuous residual GCN layers, which is the main contribution of this paper. The application and study of discrete residual learning over other meta-models has been already explored in, for example, \cite{greff2017highwayresidual,kim2017reslstm,kipf2017gcn,wang2016resrnn,zilly2017recurrenthighwaynetworks}. Furthermore, the application of continuous residual learning has been explored in \cite{chen2018odenet,gratwohl2018odenet,haber2017continuous,lu2018continuous,ruthotto2018continuous}, here we found no work on applying this technique to graph-structured data. The application of Deep Learning for learning characteristics over social networks has been studied, for example, in \cite{kipf2017gcn,li2018gcndeeper,pan2015subgraphselection,pan2013graphstream,xu2018powerfulgnn}.

Many other papers have tried to improve over the work of \cite{kipf2017gcn}. For example, \cite{xu2018powerfulgnn} shows that allowing multiple-layered convolutional kernels improve the expressivity of the GCN model, and that the neighbour aggregation method used in such a model also impacts on the number of graphs it can tell apart, proving that a sum aggregation should be preferred over a mean or max aggregation. Other work allow attentional pooling of each node's neighbours \cite{velickovic2018gat}, and also show an improvement in performance. In \cite{hamilton2017graphsage}, they experiment with different aggregation/pooling functions for a GCN, and \cite{gilmer2017mpnn} uses an edge-annotated pooling in his MPNN. Preliminary experiments with these models did not yield promising results and thus we left them for future work, focusing here on the canonical GCN model \cite{kipf2017gcn} as our baseline.

Some models in the GNN literature also employ methods that can be seen as similar to residual connections. For example, one could interpret the LSTM and GRU modules, which are often applied in GNNs \cite{gilmer2017mpnn,li2016ggnn,selsam2018neurosat}, as providing a similar feature to residual connections \cite{greff2017highwayresidual}, since they may allow information to pass along time-steps unchanged if the network learns to do so. Also, some previously published results \cite{palm2018recurrentrelational,xu2018powerfulgnn} use many or all the layers of their GNN model being used to perform gradient descent. This in some sense also allows the gradients to reach specific parts of the network without being polluted with further transformations. These models allow many-layered networks to be effectively learned and could be seen as having a similar effect to residual modules, however this is more computationally expensive than allowing residual connections.

\section{Conclusions and Future Work}\label{sec:concfut}

The results presented here suggest the application of this technique over other problems, specially those for which the use of a many-layered GNN is mandatory. Prime examples of these are the Edge Network MPNN model \cite{gilmer2017mpnn}, the myriad implementations of the Graph Network \cite{battaglia2018gn} formalisation, and the many variations on the stateless GCN model \cite{hamilton2017graphsage,velickovic2018gat,xu2018powerfulgnn}. More specifically, we believe that Interaction Networks \cite{battaglia2016interaction} could greatly benefit from being transformed to a continuous spectrum, since the physics relations they reason about are inherently defined in a continuous space. Other model which would be scientifically interesting to explore are GNNs which reason about temporal graphs \cite{jin2019tkg,trivedi2017tkg,yan2018skeleton}, we believe our formalisation here could be expanded to allow the graph topology and labels to change over the layer-space, allowing graphs that change through a layer-continuous function.

Another venue for future work is further investigation on whether depth in relational data leads to better performance or whether there are limits to this application on graphs, some of which has already been investigated by \cite{li2018gcndeeper,xu2018powerfulgnn}. One could also consider to research whether the dynamic number of layers/time-steps enjoyed by continuous residual layers allows learning transformations in the same vein as the ones achieved in \cite{selsam2018neurosat} and \cite{palm2018recurrentrelational}, whose models' learned transformations can be expanded for more time-steps of computation than originally trained and whose results improve incrementally over function applications. We believe that the results presented in \cite{chen2018odenet,gratwohl2018odenet}, fostered by the continuity of the ODE models, are a strong indicator of this, while also being a starting point for analysing what transformations can achieve a similar non-decreasing performance improvement over more steps of computation.

Finally, another possible work would be  related to improving the computational performance of the ODE solvers used in tandem with the neural networks presented here. Chen et al. \cite{chen2018odenet} already produced outstanding results with their adjoint method, which we make use of in this paper, but the learned network still needs many more applications than may be really necessary when learning with graph-structured data, and each application considered during integration implies in the multiplication of the full adjacency matrix, which is far too costly for larger graphs. A form of applying the neighbourhood aggregation without depending on costly transformations such as matrix multiplications, sparse or dense, would surely benefit this technique in the domain of graphs.

\section*{Acknowledgements}We would like to thank NVIDIA Corporation for the Quadro GPU granted to our research group. This work is partly supported by Coordenação de Aperfeiçoamento de Pessoal de Nível Superior (CAPES) -- Finance Code 001 and by the Brazilian Research Council CNPq. We would also like to thank the Pytorch developers and the authors who made their source code available to foster reproducibility in research, and thanks also to Henrique Lemos and Rafael Baldasso Audibert for their helpful discussions and help reviewing the paper.

\bibliographystyle{plain}
\bibliography{DRGNN.bbl}

\end{document}